\documentclass[letterpaper, 10 pt, conference]{ieeeconf}  
\usepackage{graphicx}
\usepackage{amsmath,amssymb}
\usepackage{booktabs}
\usepackage{multirow}
\usepackage{array}    
\usepackage{tabularx}
\usepackage[table,dvipsnames]{xcolor}
\usepackage{siunitx}
\usepackage{caption}
\usepackage{stfloats}
\usepackage{algorithm}
\usepackage{algpseudocode}
\usepackage{hyperref}
\usepackage[dvipsnames]{xcolor}
\usepackage{tabularx} 
\newcommand{\argmin}{\operatorname*{arg\,min}}
\newcommand{\F}{\mathsf{F}}
\newcommand{\Had}{\odot}

\newcolumntype{Y}{>{\centering\arraybackslash}X}      
\newcolumntype{Z}{>{\centering\arraybackslash}p{1.8cm}} 

\IEEEoverridecommandlockouts                              

\title{\LARGE \bf
Low-Rank Prehab: Preparing Neural Networks for SVD Compression
}

\author{
Haoran Qin,
Shansita Sharma,
Ali Abbasi,
Chayne Thrash,
and Soheil Kolouri
\\[1ex]
Department of Computer Science, Vanderbilt University, TN, USA
}

\begin{document}

\maketitle
\thispagestyle{empty}
\pagestyle{empty}

\begin{abstract}
Low-rank approximation methods such as singular value decomposition (SVD) and its variants (e.g., Fisher-weighted SVD, Activation SVD) have recently emerged as effective tools for neural network compression. In this setting, decomposition acts as a “surgical” intervention, followed by fine-tuning that serves as “rehab” to recover accuracy. Inspired by prehabilitation in surgery, we introduce a pre-compression fine-tuning stage, \emph{Low-Rank Prehab}, that explicitly encourages low-rank structure in weight matrices while preserving task performance. By conditioning the model before SVD, Prehab steers weights toward spectrally compact regions of the parameter space, enabling smoother low-rank approximation and improved recovery. Experiments on large language models (LLMs) and other Transformer-based architectures, including Vision Transformers (ViTs), show that Prehab substantially reduces the immediate accuracy drop after compression and consistently improves post-finetuning performance. Across a wide range of compression ratios, our method outperforms state-of-the-art SVD-based techniques such as SVD-LLM, highlighting the importance of preparing models for compression rather than only improving the compression and recovery stages. Source code is available at \url{https://github.com/niqretnuh/PREHAB-SVD}.

\end{abstract}

\section{INTRODUCTION}
\label{sec:intro}

Recent advances in computer vision, language, and multimodal reasoning have been propelled by increasingly large Transformer-based architectures such as Vision Transformers (ViTs) and Large Language Models (LLMs). While scaling has delivered remarkable accuracy gains, it also brings heavy computational, storage, and deployment costs. Models with billions of parameters remain impractical for on-device inference or cross-agent communication. Consequently, efficient compression methods that shrink model size while preserving task fidelity are essential for real-world deployment.

Among available strategies, \textit{low-rank approximation} via Singular Value Decomposition (SVD) stands out for its simplicity and compatibility with existing architectures. By decomposing a weight matrix $W \in \mathbb{R}^{m \times n}$ into two low-rank factors $U_r\Sigma_rV_r^{\top}$, SVD reduces both parameter count and multiply–accumulate cost. However, directly truncating singular values often incurs a sharp performance drop, prompting extensive research into post-training refinements that account for loss geometry or parameter sensitivity.


Most existing SVD-based approaches operate \textit{after} training. Fisher-Weighted SVD (FWSVD)~\cite{FWSVD} and Generalized Fisher-Weighted SVD (GFWSVD)~\cite{GFWSVD} apply weighting of reconstruction error based on parameter importance using diagonal or Kronecker-factored approximation of Fisher Information Matrix (FIM). Activation-SVD (ASVD)~\cite{ASVD} and SVD-LLM~\cite{SVDLLM} improve truncation by normalizing input activations through data whitening, aligning the singular spectrum with the true loss (i.e., the activation subspaces). More recently, Dobi-SVD~\cite{DOBISVD} learns heterogeneous per-layer ranks through differentiable truncation and reconstructs compressed weights using incremental PCA (IPCA) with a memory–rank bijection for remapping. Despite their differences, all these methods share a reactive design: compression is performed on a converged model, followed by fine-tuning (“rehab”) to restore accuracy. This reactive pipeline limits attainable fidelity because the pre-compression weights were never optimized to be low-rank friendly.

\begin{figure}[t]
    \centering
    \includegraphics[width=\linewidth,trim=50 40 55 40,clip]{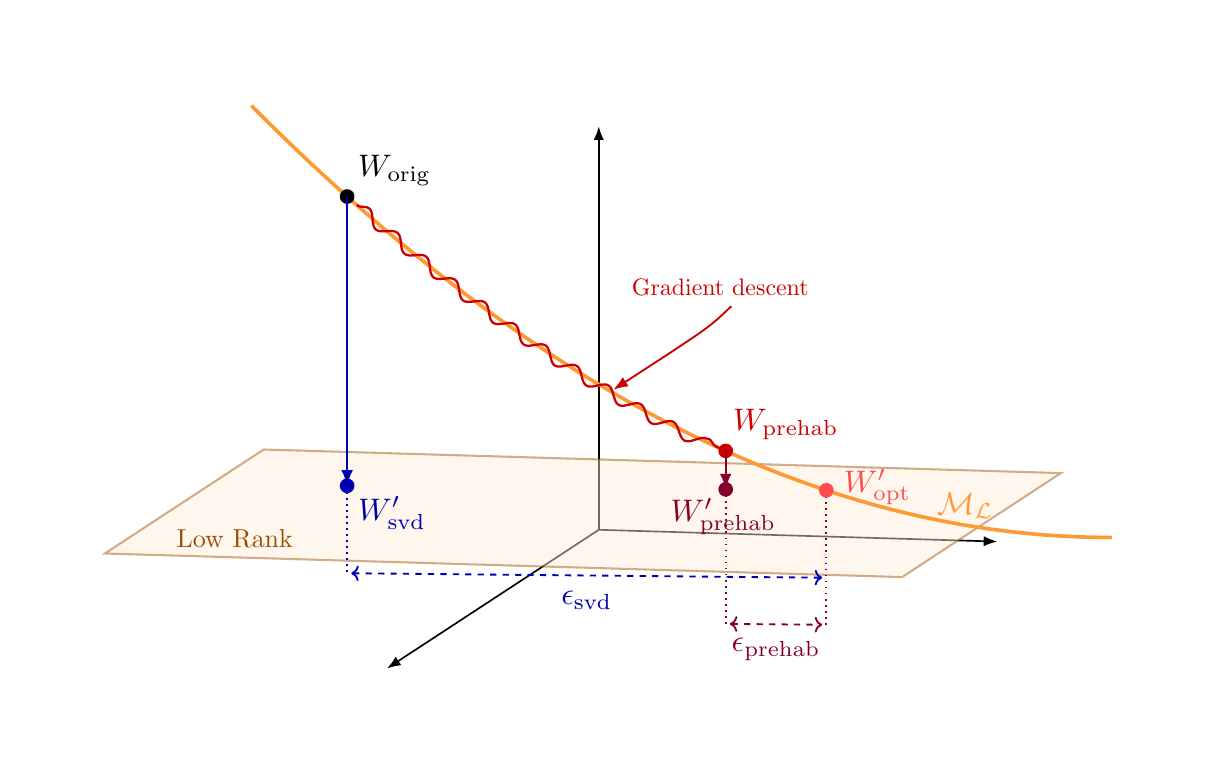}
    \vspace{-.3in}
    \caption{
    Geometric view of the proposed \textbf{Low-Rank Prehab}. 
    The original weights $W_{\mathrm{orig}}$ lie on the manifold $\mathcal{M}_{\mathcal{L}}$ of optimal solutions. 
    Standard SVD projects $W_{\mathrm{orig}}$ directly onto the low-rank plane, producing $W'_{\mathrm{SVD}}$ far from the optimal intersection $W'_{\mathrm{opt}}$.
    Prehab first moves the weights along $\mathcal{M}_{\mathcal{L}}$ via joint optimization of task loss with a smooth rank regularizer, yielding $W_{\mathrm{prehab}}$. 
    Its SVD projection $W'_{\mathrm{prehab}}$ lies closer to $W'_{\mathrm{opt}}$, reducing the geometric deviation $\epsilon_{\mathrm{prehab}}\!\ll\!\epsilon_{\mathrm{SVD}}$ and consequently minimizing post-compression loss.
    }
    \vspace{-.25in}
    \label{fig:manifold}
\end{figure}

We re-examine compression from a geometric perspective. Let $\mathcal{M}_{\mathcal{L}}$ denote the manifold of optimal solutions, i.e., weight configurations that achieve low training loss. As illustrated in Figure~\ref{fig:manifold}, standard SVD projects a converged weight matrix $W_{\mathrm{orig}}$ onto a low-rank subspace, but the resulting point $W'_{\mathrm{SVD}}$ may lie far from $\mathcal{M}_{\mathcal{L}}$,. This misalignment explains the immediate degradation of compressed models.

Most existing approaches focus on optimizing the projection/truncation step, ignoring how the original weights, $W_{\mathrm{orig}}$, are obtained. Our key insight is that compression loss often arises because training drifts away from the manifold of low-rank, near-optimal solutions. We address this by conditioning the model \textit{before} compression, guiding its learning dynamics toward regions of the loss manifold, $\mathcal{M}_{\mathcal{L}}$, that are inherently low-rank. This encourages weights that naturally align with these directions to be more robust to truncation, reducing the gap between original and compressed models.

We propose \textbf{Low-Rank Prehab}, a pre-compression fine-tuning stage that prepares the model for subsequent SVD-based compression. Drawing analogy to medicine, SVD acts as the ``surgery,'' post-compression fine-tuning as ``rehab,'' and our pre-conditioning phase as ``prehab.''  Prehab co-optimizes task loss with an auxiliary term for smooth rank surrogate, gradually steering weights along $\mathcal{M}_{\mathcal{L}}$ toward low-rank configurations while maintaining task performance.

\noindent\textbf{Contributions.} This work makes three main contributions:
\begin{enumerate}
    \item \textbf{Pre-compression conditioning framework.} We introduce \textbf{Low-Rank Prehab}, a lightweight, architecture-agnostic stage that shapes the model’s geometry for smoother SVD compression.
    \item \textbf{Geometric interpretation.} We formalize model compression as projection from the manifold of high-performing solutions onto a low-rank subspace and show that pre-conditioning reduces this manifold–subspace gap.
    \item \textbf{Empirical validation.} Experiments on Llama, ViT, and BERT demonstrate that Prehab reduces immediate post-compression accuracy loss and improves final recovery, outperforming SVD-LLM across compression ratios.
\end{enumerate}

\section{RELATED WORK}

\subsection{Fisher-Aware SVD Methods}

\textbf{Fisher-Weighted SVD (FWSVD)} replaces the Frobenius reconstruction norm with a Fisher-weighted metric, emphasizing parameters with high task sensitivity. A diagonal Fisher approximation, estimated from gradients on a calibration set, provides importance weighting but neglects correlations.  
\textbf{Generalized Fisher-Weighted SVD (GFWSVD)} extends this formulation using a Kronecker-factored (K-FAC) Fisher approximation, preserving structured dependencies between input and output dimensions. GFWSVD achieves higher fidelity under aggressive compression (up to 20×) but, like FWSVD, acts purely after convergence and cannot promote compressibility during training.

\subsection{Truncation- and Loss-Aligned SVD Methods}

\textbf{SVD-LLM} links compression loss to the singular spectrum of each layer. It first whitens activations via a Cholesky factor of their covariance so that singular values directly reflect expected loss contributions, i.e., the compression objective is aligned with the task loss. SVD is then applied in the whitened space, followed by a closed-form layerwise correction or optional LoRA fine-tuning.

\textbf{Dobi-SVD} advances post-training SVD through \textit{heterogeneous rank selection}, \textit{incremental PCA (IPCA) recovery}, and \textit{post-decomposition remapping}. Instead of a uniform rank, each layer learns its optimal compression ratio via differentiable truncation stabilized by truncated Taylor expansions. Compressed weights are reconstructed analytically using Eckart–Young–Mirsky–optimal updates with IPCA, while a memory–rank bijection ``remapping" ensures information-preserving remapping under strict storage constraints. Despite strong empirical results, Dobi-SVD remains a post-training method and does not alter the model’s pre-compression geometry.

\subsection{From Post-Projection to Pre-Conditioning}

All current SVD-based approaches focus on improving the “surgery’’—the decomposition and post-hoc recovery—yet leave the model unprepared for compression. \textbf{Low-Rank Prehab} fills this gap by conditioning the model before compression through a brief optimization phase that jointly minimizes task loss and a rank surrogate. This shifts weights toward low-rank-compatible regions of $\mathcal{M}_{\mathcal{L}}$, yielding smaller projection error and greater stability after truncation. Prehab is orthogonal to Fisher- or truncation-aware techniques and can serve as a universal pre-processing stage preceding any SVD variant. By introducing this ``pre-surgical'' preparation, we show that preparing a model for compression can be as crucial as the compression itself.

\begin{table}[t!]
\centering
\renewcommand{\arraystretch}{1.25}
\setlength{\tabcolsep}{6pt}
\captionsetup{justification=raggedright,singlelinecheck=false}
\caption{\footnotesize Core objectives for SVD-based compression methods in related work. Here, $F$ denotes the diagonal Fisher information matrix used in FWSVD; 
$S = X^{\top}X$ is the activation covariance matrix (whitening in SVD-LLM); 
and $F \approx G \otimes A$ is the Kronecker-factored Fisher approximation in GFWSVD.
$A_i = x_i W$, where $V_{A_i}$ are the right singular vectors of $A_i$, and 
$G_k = \operatorname{diag}(\mathbf{1}_k, \mathbf{0}_{d-k})$.}
\label{tab:svd_objectives}
\begin{tabularx}{\linewidth}{@{} l X @{}}
\toprule
\textbf{Method} & \textbf{Optimization (rank-$r$ approximation)} \\
\midrule

\textbf{SVD} &
\makebox[\linewidth][l]{$\displaystyle
\widehat{W}_{\text{SVD}}
= \argmin_{\operatorname{rank}(\widehat{W}) \le r}
\;\big\|\, W - \widehat{W} \,\big\|_{\F}^{2}
\;=\; U_r \Sigma_r V_r^{\top}
$} \\[0.6em]

\textbf{FWSVD} &
\makebox[\linewidth][l]{$\displaystyle
\widehat{W}_{\text{FWSVD}}
= \argmin_{\operatorname{rank}(\widehat{W}) \le r}
\;\big\|\, F^{1/2} \Had (W - \widehat{W}) \,\big\|_{\F}^{2}
$} \\[0.6em]

\textbf{SVD-LLM} &
\makebox[\linewidth][l]{$\displaystyle
\widehat{W}_{\text{SVD-LLM}}
= \argmin_{\operatorname{rank}(\widehat{W}) \le r}
\;\big\|\, (W - \widehat{W})\, S^{1/2} \,\big\|_{\F}^{2}
$} \\[0.6em]

\textbf{GFWSVD} &
\makebox[\linewidth][l]{$\displaystyle
\widehat{W}_{\text{GFWSVD}}
= \argmin_{\operatorname{rank}(\widehat{W}) \le r}
\;\big\|\, A^{1/2}\,(W - \widehat{W})\, G^{1/2} \,\big\|_{\F}^{2}
$} \\[0.6em]

\textbf{Dobi\mbox{-}SVD} &
\makebox[\linewidth][l]{$\displaystyle
\widehat{W}_{\text{Dobi}}
= \argmin_{\operatorname{rank}(\widehat{W}) \le k}
\;\sum_{i=1}^{n}\big\|\, x_i W V_{A_i} G_k V_{A_i}^{\!\top}
- x_i \widehat{W} \,\big\|_2^2
$} \\[-0.2em]
& \quad
\textit{Differentiable truncation of activations} learns $k$; 
\textit{IPCA} provides an optimal Eckart--Young--Mirsky recovery of $\widehat{W}$.\\
\bottomrule
\end{tabularx}
\end{table}

\section{METHOD}
\label{sec:method}

Previous SVD-based methods treat each trained weight matrix $W$ as static and project it directly onto the low-rank manifold. This purely algebraic step ignores the loss landscape, so large projection errors can induce substantial performance degradation. In contrast, and building on the geometric intuition of Section~I, we introduce \textbf{Low-Rank Prehab}, a lightweight \textit{pre-conditioning} stage applied before compression. Instead of reacting to post-SVD compression loss, Prehab proactively steers network parameters toward the low-rank manifold while keeping the loss low.

Formally, given pretrained weights $\{W_\ell\}_{\ell=1}^{L}$, the goal is to obtain perturbed weights $\{W_\ell^{\mathrm{prehab}}\}$ that (i) maintain task accuracy and (ii) lie closer to the manifold of low-rank matrices under the same Fisher-weighted geometry used in SVD-LLM. Recall that the rank of $W$ is the number of nonzero singular values,
$\mathrm{rank}(W) = \|\sigma(W)\|_0,$
where $\sigma(W)$ denotes the vector of singular values. Since the $\ell_0$ norm is non-differentiable, we optimize a smooth surrogate $\mathcal{R}_{\mathrm{rank}}(W_\ell X_\ell)$ defined over the Fisher-whitened matrices $W_\ell X_\ell$, where $X_\ell$ is the Cholesky factor of the (uncentered) activation covariance used in SVD-LLM. The overall Prehab objective combines task fidelity and rank regularization:
\begin{equation}
\mathcal{L}_{\mathrm{prehab}}
= \mathcal{L}_{\mathrm{task}}(\{W_\ell X_\ell X_\ell^{-1}\}_{\ell=1}^L)
+ \lambda \sum_{\ell=1}^{L} \mathcal{R}_{\mathrm{rank}}(W_\ell X_\ell),
\tag{2}
\end{equation}
where $\lambda > 0$ controls the trade-off between compressibility and accuracy, and layer-specific coefficients $\lambda_\ell$ may optionally be used. The factor $X_\ell X_\ell^{-1}$ preserves functional equivalence while enabling gradients to act on the Fisher-weighted representation $W_\ell X_\ell$.

\subsection{Prehab Loss Function}

The Prehab objective consists of a standard task loss, evaluated on the uncompressed model, and an auxiliary spectral regularizer applied to the whitened weights:
\begin{equation}
\mathcal{L}_{\mathrm{task}}
= \mathbb{E}_{(x,y)\sim\mathcal{D}}
    \big[\ell(f(x;\{W_\ell X_\ell X_\ell^{-1}\}_{\ell=1}^L),y)\big]. \tag{3}
\end{equation}
Here, $\mathcal{L}_{\mathrm{task}}$ measures predictive fidelity, while $\mathcal{R}_{\mathrm{rank}}$ penalizes high-rank structure in the Fisher-weighted space. In practice, we optimize $\mathcal{L}_{\mathrm{prehab}}$ for a few fine-tuning epochs using AdamW, freezing $\{X_\ell\}$ and letting gradients propagate through the round-trip $X_\ell X_\ell^{-1}$ transformation.

\subsection{Rank Surrogates: $\ell_1$-Norm and Stable Rank}

Because $\mathrm{rank}(W)$ is not differentiable, we consider two smooth surrogates defined on the whitened matrix $W X$:

\paragraph{$\ell_1$-norm of singular values.}
\begin{equation}
\mathcal{R}_{\ell_1}(W X)
= \sum_i |\sigma_i(W X)|, \tag{4}
\end{equation}
which imposes an $\ell_1$ penalty on the singular spectrum, promoting sparsity but requiring explicit SVD and scaling poorly for large models.

\paragraph{Stable rank \cite{StableRank}}
\begin{equation}
\mathrm{SRank}(W X)
= \frac{\|W X\|_*^{\,2}}{\|W X\|_F^{\,2}}, \tag{5}
\end{equation}
a smooth, scale-invariant proxy for true rank that depends only on matrix norms and can be efficiently approximated using randomized trace estimators or sketching methods, often in nearly linear time in the matrix size. In our experiments, we rely on such approximations, making stable-rank regularization practical even for large models. Its gradient with respect to $W$ is
\begin{equation}
\resizebox{\linewidth}{!}{$
\nabla_W \mathrm{SRank}(W X)
= 2(WX)\!\left(
\frac{\Sigma_X}{\|W X\|_F^2}
- \mathrm{SRank}(W X)\,I
\right)X^{\top},
$}
\tag{6}
\end{equation}
where $\Sigma_X$ contains the singular values of $W X$. Minimizing $\mathrm{SRank}(W X)$ directly shapes the Fisher-weighted spectrum that SVD-LLM will later truncate.

\subsection{Tunable Task–Compression Trade-Off via \texorpdfstring{$\lambda$}{lambda}}

The hyperparameter $\lambda$ governs the trade-off between accuracy and compressibility. Small $\lambda$ values preserve task fidelity, while larger ones promote stronger low-rank alignment by enhancing singular-value decay in $W X$. Empirically, logarithmic sweeps of $\lambda$ produce stable post-compression results. From a geometric standpoint, $\lambda$ controls how far Prehab moves $W$ along the solution manifold $\mathcal{M}_{\mathcal{L}}$ toward the Fisher-weighted low-rank manifold $\mathcal{M}_r^X$.

\subsection{Algorithmic Procedure}

The complete Prehab process is outlined in Algorithm~\ref{alg:prehab}. It alternates between minimizing task loss and applying Fisher-weighted rank regularization. After pre-conditioning, SVD-LLM truncation operates on a smoother spectrum with minimal degradation.

\begin{algorithm}[t]
\caption{Fisher-Aligned Low-Rank Prehab}
\label{alg:prehab}
\begin{algorithmic}[1]
\Require Pretrained weights $\{W_\ell^{(0)}\}$, whitening matrices $\{X_\ell\}$, dataset $\mathcal{D}$,
         rank coefficient $\lambda$, epochs $T_{\mathrm{prehab}}$, learning rate $\eta$
\Procedure{PreProcessing (Prehab)}{}
\For{$t = 1$ \textbf{to} $T_{\mathrm{prehab}}$}
    \State Sample minibatch $(x,y)\!\sim\!\mathcal{D}$
    \State $\mathcal{L}_{\mathrm{task}}\!\gets\!\ell(f(x;\{W_\ell^{(t)} X_\ell X_\ell^{-1}\}_{\ell=1}^{L}),y)$
    \State $\mathcal{R}_{\mathrm{rank}}\!\gets\!\sum_{\ell}\mathrm{SRank}(W_\ell^{(t)} X_\ell)$
    \State $W^{(t+1)}\!\leftarrow\!W^{(t)}-\eta\nabla_{W^{(t)}}\big(\mathcal{L}_{\mathrm{task}}
    +\lambda\,\mathcal{R}_{\mathrm{rank}}\big)$
\EndFor
\EndProcedure
\Procedure{PostProcessing (Surgery \& Rehab)}{}
    \State \textbf{Surgery:} SVD-based compression (e.g., SVD-LLM)
    \State \textbf{Rehab:} Fine-tune to recover residual loss
\EndProcedure
\end{algorithmic}
\end{algorithm}

\subsection{Summary of the Prehab Framework}

Low-Rank Prehab serves as a Fisher-aligned pre-conditioning mechanism that harmonizes training geometry with compression geometry. By minimizing the stable rank of $W X$ while preserving the functional mapping $f(x; W X X^{-1})$, it teaches the model to inhabit a subspace where subsequent SVD truncation naturally aligns with task-relevant directions. The procedure requires no architectural change, adds negligible cost, and integrates seamlessly with post-training compressors such as SVD-LLM.

\section{EXPERIMENTS}

\begin{figure}[b]
    \centering
    \vspace{-.15in}
    \includegraphics[width=0.95\linewidth]{ViT_gains_2.pdf}
    \vspace{-.05in}
    \caption{\textbf{Relative improvement of Prehab-SVD over SVD-LLM on ViT-B.} 
    Gains are computed as $(\text{Prehab} - \text{SVD-LLM}) / \text{SVD-LLM}$ for each compression rate.
    Performance gains increase with compression, indicating enhanced robustness of Prehab-SVD at higher compression rates.}
    \label{fig:vit_gains}
\end{figure}

\subsection{Vision Transformer Evaluation}

We evaluate Prehab-SVD on the ViT-B/16 \cite{ViT} architecture to examine its applicability beyond language models.
Training follows the standard ImageNet-1K \cite{Imagenet} setup with $\ell_1$-norm regularization on the singular values and cross-entropy loss. 
Each model is fine-tuned for 500 steps with batch size 64 and $\lambda = 10^{-1}$. 
For LoRA fine-tuning, we follow the SVD-LLM protocol of alternating $U$, $V$ matrix updates using LoRA of rank $r\!=\!10$, 
batch size 64, and 100 steps per factor.

\begin{table}[t]
\centering
\footnotesize
\setlength{\tabcolsep}{9pt}
\renewcommand{\arraystretch}{1.05}
\caption{ImageNet Top-1 (\%) for ViT-B across truncation levels (higher is better).}
\begin{tabular}{@{}llc@{}}
\toprule
\textbf{Ratio} & \textbf{Method} & \textbf{Top-1 Accuracy (\%)} \\
\midrule
\textbf{0\% (Baseline)} & Original ViT-B & 81.10 \\
\midrule

\multirow{7}{*}{\shortstack{20\%}}
  & SVD                 & 70.24 \\
  & FWSVD               & 71.90 \\
  & GFWSVD              & 76.75 \\
\cmidrule(lr){2-3}
  & SVD-LLM             & 76.56 \\
  & Prehab-SVD          & \textbf{78.49} \textcolor{ForestGreen}{($\uparrow$1.93)} \\
\cmidrule(lr){2-3}
  & SVD-LLM$^{\dagger}$ & 77.51 \\
  & Prehab-SVD$^{\dagger}$ & \textbf{78.68} \textcolor{ForestGreen}{($\uparrow$1.17)} \\
\midrule

\multirow{7}{*}{\shortstack{40\%}}
  & SVD                 & 39.60 \\
  & FWSVD               & 53.93 \\
  & GFWSVD              & 68.95 \\
\cmidrule(lr){2-3}
  & SVD-LLM             & 67.83 \\
  & Prehab-SVD          & \textbf{71.92} \textcolor{ForestGreen}{($\uparrow$4.09)} \\
\cmidrule(lr){2-3}
  & SVD-LLM$^{\dagger}$ & 72.95 \\
  & Prehab-SVD$^{\dagger}$ & \textbf{74.41} \textcolor{ForestGreen}{($\uparrow$1.46)} \\
\midrule

\multirow{7}{*}{\shortstack{50\%}}
  & SVD                 & 10.06 \\
  & FWSVD               & 32.61 \\
  & GFWSVD              & 60.48 \\
\cmidrule(lr){2-3}
  & SVD-LLM             & 58.42 \\
  & Prehab-SVD          & \textbf{64.06} \textcolor{ForestGreen}{($\uparrow$5.64)} \\
\cmidrule(lr){2-3}
  & SVD-LLM$^{\dagger}$ & 66.41 \\
  & Prehab-SVD$^{\dagger}$ & \textbf{68.59} \textcolor{ForestGreen}{($\uparrow$2.18)} \\
\midrule

\multirow{7}{*}{\shortstack{60\%}}
  & SVD                 & 0.52 \\
  & FWSVD               & 11.95 \\
  & GFWSVD              & 43.18 \\
\cmidrule(lr){2-3}
  & SVD-LLM             & 28.38 \\
  & Prehab-SVD          & \textbf{48.12} \textcolor{ForestGreen}{($\uparrow$19.74)} \\
\cmidrule(lr){2-3}
  & SVD-LLM$^{\dagger}$ & 50.06 \\
  & Prehab-SVD$^{\dagger}$ & \textbf{58.97} \textcolor{ForestGreen}{($\uparrow$8.91)} \\
\bottomrule
\end{tabular}

\vspace{3pt}
\footnotesize
\raggedright
{$^{\dagger}$\,Lightweight LoRA fine-tuning (“+LoRA”). 
Green $\uparrow$ indicates Top-1 accuracy improvement vs.\ SVD-LLM at the same ratio. 
“Ratio” = fraction of parameters removed in compressed blocks; 
Groups: \emph{Baselines} (SVD/FWSVD/GFWSVD), \emph{Loss-aligned} (SVD-LLM, Prehab-SVD), and \emph{+LoRA variants}. }
\label{tab:vitb_preha_svd}
\vspace{-.3in}
\end{table}

\vspace{4pt}
\noindent\textbf{Results without LoRA.}
Table~\ref{tab:vitb_preha_svd} shows the Top-1 accuracy of ViT-B under varying compression ratios.
Prehab-SVD consistently improves over SVD-LLM across all compression levels, demonstrating its effectiveness as a spectral pre-conditioning step.
At moderate, 50\% compression, Prehab-SVD improves accuracy from 58.42\% (SVD-LLM) to 64.06\%, yielding a +5.64\% improvement.
The benefit becomes more pronounced as compression increases: at 60\% compression, Prehab-SVD reaches 48.12\% accuracy compared to 28.38\% for SVD-LLM, corresponding to a substantial +19.74\% improvement.
These results show that pre-conditioning becomes stronger in higher spectral sparsity regimes.

\vspace{4pt}
\noindent\textbf{Results with LoRA fine-tuning.}
When applying the lightweight LoRA fine-tuning stage (denoted by $^{\dagger}$), Prehab-SVD retains its advantage across all configurations.
Although the relative gains narrow at lower compression levels, Prehab-SVD$^{\dagger}$ continues to outperform SVD-LLM$^{\dagger}$.
At 50\% compression, Prehab-SVD$^{\dagger}$ improves accuracy from 66.41\% (SVD-LLM$^{\dagger}$) to 68.59\%, yielding a +2.18\% gain.
At 60\% compression, Prehab-SVD$^{\dagger}$ reaches 58.97\%, compared to 50.06\% for SVD-LLM$^{\dagger}$, achieving a +8.61\% improvement.
This indicates that pre-conditioning helps LoRA recover lost performance more effectively, yielding better convergence and spectral alignment.

\vspace{4pt}
\noindent\textbf{Prehab-SVD significantly improves performance at high compression rates.}
Figure~\ref{fig:vit_gains} reports the percentage gain of Prehab-SVD over SVD-LLM, computed as 
$(\text{Prehab} - \text{SVD-LLM}) / \text{SVD-LLM}$ to reveal the true magnitude of improvement. 
We observe that performance gains grow exponentially with compression rate, reflecting improved robustness to aggressive compression and highlighting the importance of \emph{prehab} at such magnitudes of compression.

\vspace{4pt}
\noindent\textbf{$\lambda$ Ablation.}
We further validate the effect of differing $\lambda$ on ViT performance. To investigate this, we set $\lambda = 10^{-1}$, $\lambda = 5.0$, and $\lambda = 10.0$ for multiple compression ratios and showcase the results in Table~\ref{tab:vitb_ablation}. We observe that $\lambda = 10.0$ notably enhances performance under the highest compression, while introducing trade-offs at lower compression ratios. 
This behavior reflects the importance of identifying a \emph{sweet spot} for the prehab regularization strength $\lambda$, balancing compression-target alignment and task fidelity for optimal performance.

\noindent\textbf{Runtime.}
All ViT-B experiments were trained on a single NVIDIA A6000 GPU. 
The Prehab stage required only 7.5 minutes in total, averaging 0.9 seconds per step due to the small additional computational overhead of the rank-regularization term, confirming that Prehab introduces negligible additional cost relative to baseline SVD-based pipelines while achieving significant gains in performance.

\subsection{BERT-Base Evaluation}
\begin{table*}[t]
\centering
\footnotesize
\setlength{\tabcolsep}{4pt} 
\renewcommand{\arraystretch}{1.0}
\caption{Performance of BERT-Base on GLUE tasks across truncation levels. Best results per block are in \textbf{bold}.}
\vspace{-.1in}\begin{tabularx}{\textwidth}{@{}clYYYYYYYZZ@{}}
\toprule
\textbf{Ratio} & \textbf{Method} & \textbf{CoLA} & \textbf{SST-2} & \textbf{MRPC} & \textbf{RTE} & \textbf{STS-B} & \textbf{QNLI} & \textbf{QQP} & \textbf{MNLI-m / MNLI-mm} & \textbf{Average} \\
\midrule
0\% & BERT-Base & 0.567 & 0.925 & 0.898 & 0.658 & 0.857 & 0.903 & 0.876 & 0.827 / 0.832 & 0.816 \\
\midrule

\multirow{5}{*}{20\%}
 & SVD & 0.021 & 0.841 & 0.684 & 0.520 & 0.616 & 0.717 & 0.727 & 0.579 / 0.583 & 0.588 \\
 & FWSVD & 0.287 & 0.890 & 0.825 & 0.552 & 0.821 & 0.862 & 0.838 & 0.705 / 0.698 & 0.722 \\
 & GFWSVD & 0.399 & 0.888 & 0.855 & 0.523 & 0.644 & 0.828 & 0.828 & 0.718 / 0.723 & 0.711 \\
 \cmidrule(lr){2-11}
 & SVD-LLM & 0.452 & 0.898 & 0.884 & 0.527 & 0.847 & 0.871 & 0.847 & 0.808 / 0.812 & 0.767 \\
 & Prehab-SVD & \textbf{0.492} & \textbf{0.914} & \textbf{0.901} & \textbf{0.614} & \textbf{0.852} & \textbf{0.876} & \textbf{0.853} & \textbf{0.812 / 0.818} & \textbf{0.790} \textcolor{ForestGreen}{($\uparrow$0.023)} \\
\midrule

\multirow{5}{*}{40\%}
 & SVD & 0.000 & 0.789 & 0.075 & 0.509 & 0.437 & 0.645 & 0.596 & 0.451 / 0.456 & 0.438 \\
 & FWSVD & 0.000 & 0.828 & 0.823 & 0.502 & 0.810 & 0.817 & 0.801 & 0.546 / 0.549 & 0.641 \\
 & GFWSVD & 0.219 & 0.852 & 0.724 & 0.462 & 0.601 & 0.809 & 0.809 & 0.632 / 0.633 & 0.638 \\
 \cmidrule(lr){2-11}
 & SVD-LLM & 0.410 & 0.891 & 0.869 & 0.486 & 0.809 & 0.784 & 0.829 & 0.781 / 0.772 & 0.732 \\
 & Prehab-SVD & \textbf{0.421} & \textbf{0.905} & \textbf{0.887} & \textbf{0.581} & \textbf{0.821} & \textbf{0.818} & \textbf{0.838} & \textbf{0.790 / 0.789} & \textbf{0.758} \textcolor{ForestGreen}{($\uparrow$0.026)} \\
\midrule

\multirow{5}{*}{60\%}
 & SVD & 0.000 & 0.627 & 0.000 & 0.523 & 0.114 & 0.561 & 0.498 & 0.356 / 0.352 & 0.335 \\
 & FWSVD & 0.000 & 0.799 & 0.811 & 0.516 & 0.680 & 0.667 & 0.710 & 0.445 / 0.447 & 0.584 \\
 & GFWSVD & 0.072 & 0.737 & 0.007 & 0.469 & 0.677 & 0.670 & 0.680 & 0.472 / 0.485 & 0.474 \\
 \cmidrule(lr){2-11}
 & SVD-LLM & 0.259 & 0.836 & 0.830 & 0.520 & 0.662 & 0.651 & 0.763 & 0.658 / 0.659 & 0.647 \\
 & Prehab-SVD & \textbf{0.309} & \textbf{0.877} & \textbf{0.847} & \textbf{0.567} & \textbf{0.681} & \textbf{0.671} & \textbf{0.771} & \textbf{0.698 / 0.695} & \textbf{0.677} \textcolor{ForestGreen}{($\uparrow$0.030)} \\
\bottomrule
\end{tabularx}

\vspace{3pt}
\footnotesize
We report F1 for QQP/MRPC, Spearman for STS-B, and Acc. for the others; green $\uparrow$ marks performance relative to SVD-LLM at the same compression.
\vspace{-.25in}
\label{tab:glue_bert}
\end{table*}

We evaluate Prehab-SVD on BERT-Base \cite{BERT} over the GLUE benchmark \cite{GLUE} to examine its effectiveness on smaller transformer models for natural language understanding. For each GLUE task, prior to applying any compression, the BERT-Base model was finetuned on the corresponding dataset for 3 epochs with a cosine scheduler with an initial learning rate $5 \times 10^{-5}$ for CoLA, and $2 \times 10^{-5}$ for the remaining tasks. Prehab was applied as a short pre-conditioning phase prior to SVD-LLM based compression for 1 epoch with batch size 64 and learning rates identical to the initial finetuning phase, jointly optimizing cross-entropy loss with the stable-rank regularizer with regularization coefficient $\lambda = 10^{-1}$.

\vspace{4pt}
\noindent\textbf{Effect without LoRA Fine-tuning.} Table~\ref{tab:glue_bert} presents the performance gain of applying Prehab prior to compression under different compression ratios. Without any LoRA adaptation, Prehab-SVD consistently improves over the post-training baseline across most scenarios, confirming that pre-conditioning enhances low-rank compatibility and reduces performance loss after compression. Notably, our method yields substantial gains at 60\% compression, especially on tasks such as RTE and CoLA, relative to the baseline SVD-LLM that we build upon. On average, Prehab-SVD offers strong performance gains across compression rates on BERT.

\noindent\textbf{Effect with LoRA Fine-tuning.} After applying lightweight LoRA recovery, the benefits of Prehab-SVD become less pronounced on BERT, as LoRA’s parameter-efficient updates largely replicate prehab's low-rank regularization effects, reducing the relative impact of pre-conditioning. Results for this setting are omitted due to space limitations. 

\noindent\textbf{Runtime.} All BERT experiments were trained on a single NVIDIA A6000 GPU. The Prehab stage was applied with the fast stable rank surrogate and required varying durations dependent on the size of the dataset, at the average rate of 1.1 seconds for each batch, confirming the negligible overhead.

\subsection{LLaMA-7B Evaluation}
We further evaluate \textbf{Prehab-SVD} on \textbf{LLaMA-7B} to assess its effectiveness on large-scale transformer architectures. 
Following the experimental design of SVD-LLM, all methods are tested under identical compression ratios on LlaMa-7B \cite{llama} and evaluated on both \textbf{WikiText-2} \cite{WikiText2} and \textbf{C4} \cite{C4}.
Prehab was applied as a short pre-conditioning phase using $\lambda = 10^{-3}$ and learning rate $10^{-5}$, jointly optimizing cross-entropy loss and the stable-rank regularizer for $3$ epochs. 
Each prehab run used $256$ WikiText-2 samples; LoRA fine-tuning (when applicable) was performed on $50\text{k}$ Alpaca \cite{alpaca} samples.

\vspace{4pt}
\noindent\textbf{Effect without LoRA Fine-tuning.}
Table~\ref{tab:ours_wiki_c4} summarizes the performance of various SVD-based compression algorithms across compression ratios. 
Without any LoRA adaptation, Prehab-SVD consistently improves over all post-training baselines on WikiText-2.
At $20\%$ compression, Prehab-SVD achieves an $11.5\%$ perplexity reduction versus SVD-LLM (7.03 vs.\ 7.94), and at $60\%$ compression, a $29.8\%$ reduction (46.75 vs.\ 66.62). 
However, on the larger and more diverse C4 dataset, Prehab-SVD without LoRA shows mild overfitting to WikiText-2, with degradation at lower compression levels (e.g., $+131.3\%$ at $20\%$ ratio).

\vspace{4pt}
\noindent\textbf{Effect with Lightweight LoRA Recovery.}
After applying a lightweight LoRA fine-tuning (“rehab”) stage on Alpaca, Prehab-SVD preserves its performance advantage while regaining generalization on C4.
For instance, Prehab-SVD$^{\dagger}$ improves over SVD-LLM$^{\dagger}$ by $6.3\%$ on WikiText-2 and $10.1\%$ on C4 at $20\%$ compression, and by $12.1\%$ and $10.1\%$ respectively at $60\%$. 
These results demonstrate that pre-conditioning yields benefits beyond immediate compression fidelity, enabling faster and more stable recovery during downstream adaptation.

\vspace{4pt}
\noindent\textbf{Summary.}
Across all ratios, Prehab-SVD narrows the perplexity gap between compressed and original models while improving robustness to data distribution shift. 
The combination of spectral pre-conditioning and lightweight LoRA recovery achieves state-of-the-art efficiency–performance trade-offs for large-scale low-rank adaptation.

\begin{table}[t]
\footnotesize
\setlength{\tabcolsep}{9pt}
\renewcommand{\arraystretch}{1.05}
\caption{\footnotesize
Perplexity ($\downarrow$) of LLaMA-7B compressed by various SVD-based methods (lower is better). 
}
\vspace{-.1in}
\centering
\begin{tabular}{@{}llcc@{}}
\toprule
\textbf{Ratio} & \textbf{Method} & \textbf{WikiText-2 $\downarrow$} & \textbf{C4 $\downarrow$} \\
\midrule
\textbf{0\%} & Original & 5.68 & 7.34 \\
\midrule

\multirow{7}{*}{\shortstack{20\%\\(10.2\,GB)}}
  & SVD                  & 20061  & 18800  \\
  & FWSVD                & 1727   & 1511   \\
  & ASVD                 & 11.14  & 15.93  \\
\cmidrule(lr){2-4}
  & SVD-LLM              & 7.94   & 15.84  \\
  & Prehab-SVD           & \textbf{7.03} \textcolor{ForestGreen}{($\downarrow$11.5\%)} & 36.64 \textcolor{Red}{($\uparrow$131.3\%)} \\
\cmidrule(lr){2-4}
  & SVD-LLM$^{\dagger}$  & 7.73   & 12.23  \\
  & Prehab-SVD$^{\dagger}$ & 7.24 \textcolor{ForestGreen}{($\downarrow$6.3\%)} & \textbf{10.99} \textcolor{ForestGreen}{($\downarrow$10.1\%)} \\
\midrule

\multirow{7}{*}{\shortstack{40\%\\(7.76\,GB)}}
  & SVD                  & 52489  & 47774  \\
  & FWSVD                & 18156  & 12847  \\
  & ASVD                 & 1407   & 1109   \\
\cmidrule(lr){2-4}
  & SVD-LLM              & 13.73  & 75.42  \\
  & Prehab-SVD           & 11.12 \textcolor{ForestGreen}{($\downarrow$19.0\%)} & 118.24 \textcolor{Red}{($\uparrow$56.7\%)} \\
\cmidrule(lr){2-4}
  & SVD-LLM$^{\dagger}$  & 9.27   & 15.63  \\
  & Prehab-SVD$^{\dagger}$ & \textbf{9.15} \textcolor{ForestGreen}{($\downarrow$1.3\%)} & \textbf{13.67} \textcolor{ForestGreen}{($\downarrow$12.5\%)} \\
\midrule

\multirow{7}{*}{\shortstack{60\%\\(5.35\,GB)}}
  & SVD                  & 105474 & 106976 \\
  & FWSVD                & 32194  & 29292  \\
  & ASVD                 & 57057  & 43036  \\
\cmidrule(lr){2-4}
  & SVD-LLM              & 66.62  & 471.83 \\
  & Prehab-SVD           & 46.75 \textcolor{ForestGreen}{($\downarrow$29.8\%)} & 298.59 \textcolor{ForestGreen}{($\downarrow$36.7\%)} \\
\cmidrule(lr){2-4}
  & SVD-LLM$^{\dagger}$  & 15.00  & 26.26  \\
  & Prehab-SVD$^{\dagger}$ & \textbf{13.19} \textcolor{ForestGreen}{($\downarrow$12.1\%)} & \textbf{23.60} \textcolor{ForestGreen}{($\downarrow$10.1\%)} \\
\bottomrule
\end{tabular}
\footnotesize
\raggedright
$^{\dagger}$\,Indicates lightweight LoRA fine-tuning (“+LoRA”).  
Green $\downarrow$ and red $\uparrow$ denote relative perplexity change vs.\ SVD-LLM at the same ratio, computed as $(\text{Prehab} - \text{SVD-LLM}) / \text{SVD-LLM}$.  
Groups: \emph{Baselines} (SVD/FWSVD/ASVD), \emph{Loss-aligned} (SVD-LLM, Prehab-SVD), and \emph{+LoRA variants}. 
\vspace{-.3in}
\label{tab:ours_wiki_c4}
\end{table}

\vspace{4pt}
\noindent\textbf{Runtime and Computational Cost.}
All experiments were conducted on a single NVIDIA H200 GPU.
The \textit{Prehab} stage was lightweight, requiring only one epoch of fine-tuning on WikiText-2, which completed in approximately 10 minutes. 
LoRA fine-tuning on the Alpaca dataset required 30 minutes per epoch and was trained for 6 epochs in total—three epochs each for the $U$ and $V$ projection matrices—resulting in an overall runtime of roughly 3 hours. 
Compared to the multi-day fine-tuning schedules typical for large-scale compression pipelines, this demonstrates that Prehab introduces minimal overhead while providing substantial compression readiness and downstream generalization improvements.

\section{CONCLUSION}
We introduced \emph{Low-Rank Prehab}, a lightweight pre-conditioning stage that aligns training and compression geometries by minimizing a smooth, loss-aligned stable-rank surrogate on activation-weighted parameter weights while preserving task loss. Across ViT-B/16 (ImageNet-1K), BERT-Base (GLUE), and LLaMA-7B (WikiText-2/C4), Prehab consistently improves performance—with or without subsequent LoRA—at negligible additional cost.

\noindent\textbf{Future work.} 
Our results suggest several natural extensions:
(i) \emph{Heterogeneous-rank Prehab:} extending the framework to layer-specific rank targets as in Dobi-SVD, where $\lambda_\ell$ adapts to each layer’s curvature or Fisher energy. This would couple Prehab with per-layer compression planning and better exploit cross-layer redundancy.
(ii) \emph{Adaptive regularization schedules:} learning $\lambda$ through bilevel or meta-optimization that directly minimizes post-compression validation loss, enabling automatic trade-off discovery between compressibility and task fidelity.
(iii) \emph{Unified pre-conditioning across modalities:} exploring Prehab’s compatibility with other efficiency paradigms such as quantization, pruning, and KV-cache compression, where shared low-rank alignment could jointly stabilize multiple compression operators.

\section*{Acknowledgment}
This work was supported by the NSF CAREER Award No. 2339898 and by generous cloud computing provided by the Lambda Research Grant.



\begin{table}[t]
\centering
\footnotesize
\setlength{\tabcolsep}{9pt}
\renewcommand{\arraystretch}{1.05}
\caption{\textbf{ImageNet Top-1 (\%) for ViT-B/16 across truncation levels (higher is better).}}
\begin{tabular}{@{}llc@{}}
\toprule
\textbf{Ratio} & \textbf{Method} & \textbf{Top-1 Accuracy (\%)} \\
\midrule
\textbf{0\% (Baseline)} & Original ViT-B/16 & 81.10 \\
\midrule

\multirow{3}{*}{\shortstack{40\%}}
  & SVD-LLM                     & 67.83 \\
  & Prehab-SVD ($\lambda{=}10^{-1}$) & \textbf{71.92} \textcolor{ForestGreen}{($\uparrow$4.09)} \\
  & Prehab-SVD ($\lambda{=}5.0$)     & 71.76 \textcolor{ForestGreen}{($\uparrow$3.93)} \\
  & Prehab-SVD ($\lambda{=}10.0$)     & 55.66 \textcolor{Red}{($\downarrow$12.17)} \\
\midrule

\multirow{3}{*}{\shortstack{50\%}}
  & SVD-LLM                     & 58.42 \\
  & Prehab-SVD ($\lambda{=}10^{-1}$) & \textbf{64.06} \textcolor{ForestGreen}{($\uparrow$5.64)} \\
  & Prehab-SVD ($\lambda{=}5.0$)     & 63.72 \textcolor{ForestGreen}{($\uparrow$5.3)} \\
  & Prehab-SVD ($\lambda{=}10.0$)     & 51.61 \textcolor{Red}{($\downarrow$6.81)} \\
\midrule

\multirow{3}{*}{\shortstack{60\%}}
  & SVD-LLM                     & 28.38 \\
  & Prehab-SVD ($\lambda{=}10^{-1}$) & 48.12 \textcolor{ForestGreen}{($\uparrow$19.74)} \\
  & Prehab-SVD ($\lambda{=}5.0$)     & \textbf{49.25} \textcolor{ForestGreen}{($\uparrow$20.87)} \\
  & Prehab-SVD ($\lambda{=}10.0$)     & 43.59 \textcolor{ForestGreen}{($\uparrow$15.21)} \\
\midrule

\multirow{3}{*}{\shortstack{80\%}}
  & SVD-LLM                     & 0.72 \\
  & Prehab-SVD ($\lambda{=}10^{-1}$) & 2.90 \textcolor{ForestGreen}{($\uparrow$2.18)} \\
  & Prehab-SVD ($\lambda{=}5.0$)     & 2.47 \textcolor{ForestGreen}{($\uparrow$1.75)} \\
  & Prehab-SVD ($\lambda{=}10.0$)     & \textbf{13.74} \textcolor{ForestGreen}{($\uparrow$13.02)} \\
\bottomrule
\end{tabular}

\vspace{3pt}
\footnotesize
\raggedright
Green $\uparrow$/red $\downarrow$ show Top-1 accuracy change vs.\ SVD-LLM at the same truncation. 
“Ratio” = fraction of parameters removed in compressed blocks. 
\label{tab:vitb_ablation}
\vspace{-.1in}
\end{table}

\bibliographystyle{IEEEtran}
\bibliography{references}

\addtolength{\textheight}{-12cm}   

\end{document}